# Evaluating Multiple Large Language Models in pediatric ophthalmology

Jason Holmes[1]*, Rui Peng[2]*, Yiwei Li[4]*, Jinyu Hu[2,3], Zhengliang Liu[4], Zihao Wu[4], Huan Zhao[5], Xi Jiang[5], Wei Liu[1], Hong Wei[2,3], Jie Zou[2,3], Tianming Liu[4] #, Yi Shao[3]#

[1]Department of Radiation Oncology, Mayo Clinic

[2]Department of Ophthalmology, the First Affiliated Hospital of Nanchang University, Nanchang 330006, China;

[3]Department of Ophthalmology, Eye& ENT Hospital of Fudan University, shanghai 200030, China;

[4]School of Computing, University of Georgia;

[5]School of Life Science and Technology, University of Electronic Science and Technology of China;

*These authors have contributed equally to this work.

#**Address correspondence to:**

Yi Shao (Email: freebee99@163.com; Tel:+086 791-88692520, Fax:+086 791-88692520), Department of ophthalmology, Eye& ENT Hospital of Fudan University, shanghai 200030, China

**Conflict of Interest Statement:** This was not an industry supported study. The authors report no conflicts of interest in this work.

**Fund program:** National Natural Science Foundation of China (82160195); Jiangxi Double-Thousand Plan High-Level Talent Project of Science and Technology Innovation (jxsq2023201036); Key R & D Program of Jiangxi Province (20223BBH80014)

**Ethical Statement:** All research methods were approved by the committee of the medical ethics of the First Affiliated Hospital of Nanchang University and were in accordance with the 1964 Helsinki declaration and its later amendments or comparable ethical standards. All subjects were explained the purpose, method, potential risks and signed an informed consent form.

**Abstract**

IMPORTANCE The response effectiveness of different large language models (LLMs) and various individuals, including medical students, graduate students, and practicing physicians, in pediatric ophthalmology consultations, has not been clearly established yet.

OBJECTIVE Design a 100-question exam based on pediatric ophthalmology to evaluate the performance of LLMs in highly specialized scenarios and compare them with the performance of medical students and physicians at different levels.

DESIGN, SETTING, AND PARTICIPANTS This survey study assessed three LLMs, namely ChatGPT (GPT-3.5), GPT-4, and PaLM2, were assessed alongside three human cohorts: medical students, postgraduate students, and attending physicians, in their ability to answer questions related to pediatric ophthalmology. It was conducted by administering questionnaires in the form of test papers through the LLM network interface, with the valuable participation of volunteers.

MAIN OUTCOMES AND MEASURES Mean scores of LLM and humans on 100 multiple-choice questions, as well as the answer stability, correlation, and response confidence of each LLM.

RESULTS GPT-4 performed comparably to attending physicians, while ChatGPT (GPT-3.5) and PaLM2 outperformed medical students but slightly trailed behind postgraduate students. Furthermore, GPT-4 exhibited greater stability and confidence when responding to inquiries compared to ChatGPT (GPT-3.5) and PaLM2.

CONCLUSIONS AND RELEVANCE Our results underscore the potential for LLMs to provide medical assistance in pediatric ophthalmology and suggest significant capacity to guide the education of medical students.

1.Introduction

The application of large language models (LLMs) has provided significant advancement in the field of Natural Language Processing (NLP).[1] Traditional approaches in NLP have involved supervised training to create and establish basic solutions for downstream tasks, including statistical methods, machine learning methods, and rule-based methods. For instance, XLNet[2, 3] introduced an unsupervised pre-training approach that models all possible permutations to learn language representations without order restrictions. This enabled XLNet to better handle long-range dependencies and co-reference resolution. On Twitter, it has been used to identify optimistic or pessimistic sentiment by recognizing contextual cues.[4] Another NLP technique is the pre-trained model BERT, which only requires the addition of extra network layers to fine-tune labeled training data for tasks like question answering and classification.[5] In certain specific scenarios, training this model in specific domains can yield significant advantages and achieve superior performance. Currently, its applications extend to the realm of medicine, exemplified by BioBERT, a specialized NLP model pre-trained on extensive biomedical corpora. Its prowess in biomedical text mining tasks is truly exceptional.[6] Simultaneously, LLMs also have excelled in few-shot learning capabilities and large-scale processing, offering solutions to contextual NLP problems and reducing or even eliminating the need for labeled training data.[7, 8] During LLM training, the desired responses from users' descriptive inputs can be inferred,[7, 9] such as "extract keywords from the following text," which prompts the language model to extract keywords from a given paragraph. In summary, LLMs have greatly facilitated advancements in NLP and have the potential to eliminate the necessity for supervised fine-tuning. Additionally, contextual learning significantly reduces the need for expensive and time-consuming manual annotation.[7, 10] Model applications are particularly valuable in the medical and scientific domains, in which limited data is often a challenge.[11-13]

Recently, the emergence of ChatGPT, based on GPT-3, has showcased unprecedented convenience in language models compared to XLNet, BERT, and others. GPT-3 itself is a pre-trained model[7] that learns rich language knowledge and patterns through unsupervised training on a massive amount of data. With 175 billion parameters, it was one of the largest language models when initiated, enabling it to better understand and generate natural language. GPT-3 has demonstrated impressive performance in various NLP tasks, ranging from text summarization to named entity recognition. ChatGPT inherits these capabilities and leverages its vast database. Furthermore, ChatGPT is trained using Reinforcement Learning from Human Feedback (RLHF), which utilizes human expertise or feedback to accelerate the reinforcement learning process. It incorporates human feedback as supplementary signals, along with the reward signals provided by the environment, to train ChatGPT in generating outputs that are most appealing and relevant to human users. The functionality of ChatGPT supports a wide range of practical applications[14-17], such as ensuring the accuracy of clinical radiology information retrieval[18] or providing powerful writing assistance.[19]

Although the technical details of OpenAI's latest upgrade; i.e, GPT-4, have not been publicly disclosed, it has demonstrated superior performance in various application scenarios.[10, 13, 20] GPT-4 is currently being utilized to support Bing, which is Microsoft's search engine, showcasing the unparalleled potential of LLMs. It enables semantic segmentation and data fusion of 3D city and small-scale unmanned aerial vehicle photogrammetry data,[21] as well as precise translation of news and political texts.[22, 23]

GPT-4 has proven to be exceptionally outstanding in various academic and professional domains. It successfully passed all six years of the Japanese National Medical Practitioner Examination.[24] Additionally, the score rate for the Neurosurgery Board Examination using GPT-4 was 83.4%, surpassing the user score rate of 72.8% by approximately 10 percentage points.[25] It is worth noting that a LLM's superiority may not be evident in simpler or widely known topics, highlighting the importance of selecting more obscure and specialized domains when evaluating the performance of a LLM. Pediatric ophthalmology, with its specialized barriers and need for extensive expertise,[26] serves as a suitable field for fair evaluation of the performance of a LLM compared to highly accessible knowledge databases.

An important factor in evaluating the accuracy of LLMs is ensuring that the test questions are not included in the training data. To address this, in our study, we created a multiple-choice exam. We evaluated three Transformer-based LLMs: ChatGPT (GPT-3.5),[8] GPT-4,[20] and PaLM2.[27] We compared their performance with doctors, graduate students, and undergraduate students in the field of pediatric ophthalmology by determining the level of performance for each model. We also explored the stability and confidence of these three LLMs in the specialized pediatric ophthalmology test. In the future, we will continue to explore the potential of GPT-4 as a guide for undergraduate students and as an assistant in diagnosis and treatment.

**2.Related work**

2.1 LLM

Transformer-based Pre-trained Language Models (PLMs), such as BERT,[5] GPT,[28] and XLNet,[3] have revolutionized the field of NLP. They have surpassed previous methods, such as models based on RNNs and CNNs, in many tasks, enhancing performance in various aspects and igniting research interest in language models.[29] Generally, PLMs can be categorized into three types: autoregressive models (e.g., GPT), masked language models (e.g., BERT), and encoder-decoder models (e.g., BART[30] and T5[31]). Recently, a series of highly powerful language models have emerged, including GPT-3,[7] Bloom,[32] PaLM,[27] and OPT.[33] These models are rooted in the Transformer architecture and have been inspired by models like BERT and GPT, but they have been developed on a larger scale.

The objective of LLMs is to learn universal language representations through pre-training, providing powerful feature representation capabilities for various NLP tasks[29] and advancing natural language understanding and generation. For example, the interpretation of the word "tissue" can vary significantly between the medical field and general contexts. Smaller language models often require continuous pre-training and supervised fine-tuning on downstream tasks to achieve acceptable performance.[17] However, LLMs have the potential to eliminate the need for fine-tuning while maintaining competitiveness.[7]

In addition to advancements in model architecture, scale, and training strategies, a LLM can further align with human preferences through RLHF.[34] This approach has been implemented in various LLMs, such as Sparrow, InstructGPT,[35] and ChatGPT.[8] ChatGPT employs RLHF to learn from human feedback, following prompts, and generating comprehensive and specific responses. Currently, ChatGPT has become a highly successful AI chatbot, achieving human-like interactions using GPT-3.5.

RLHF incorporates human feedback into the generation and selection of optimal results,[36] training a reward model based on human preferences for ranking generated outputs. This reward model rewards outputs that align most closely with human preferences and values. This makes ChatGPT an ideal candidate model for this study.

The recent development of GPT-4 has significantly advanced the state-of-art in language modeling. GPT-4 showcases exceptional reasoning abilities, creativity, image understanding, contextual comprehension, and multimodal capabilities, resulting in more complex and diverse responses. The success of large-scale GPT models has spurred exploration into specialized models for specific domains, such as LLMs tailored for medical specialties, which has the potential to revolutionize these fields.

2.2 Language models and exams

LLMs possess exceptional natural language understanding capabilities. Additionally, they undergo training with massive amounts of data that provide extensive knowledge. These features make LLMs ideal candidates for academic and professional benchmarks.

Recently, OpenAI released their first research paper evaluating the performance of LLMs in academic and professional exams designed for educated individuals. The study assessed the capabilities of GPT-4 in various subjects, ranging from standardized law exams to the GRE. The results demonstrated that GPT-4 performed exceptionally well across these exams. Furthermore, research conducted in Japan and by Microsoft indicated that GPT-4 has a high probability of passing the United States Medical Licensing Examination (USMLE)[37] and the licensing exams for medical practitioners.[24] These findings highlight the potential of GPT-4 to excel in specialized exams, including those in the medical field.

This research represents the first evaluation of LLMs in the field of pediatric ophthalmology. We believe that it can pave the way for future studies evaluating LLMs in highly specialized branches of medicine.

## 3. Methods

This study created a pediatric ophthalmology test consisting of 100 multiple-choice questions from the question banks of the pediatric ophthalmology board exam and medical school final exams which was given by Professor of surgery in Nanchang University. The test covered questions on the following topics: pediatric cataracts (10 questions), common tumors in children (10 questions), pediatric conjunctival diseases (14 questions), pediatric lacrimal system diseases (10 questions), pediatric glaucoma (10 questions), ocular manifestations of systemic diseases in children (7 questions), amblyopia in children (11 questions), pediatric retinal disorders (9 questions), pediatric strabismus (10 questions), and pediatric ocular melanoma (10 questions). The exam questions are listed in the Appendix.

The 100 multiple-choice questions in pediatric ophthalmology were inputted into each LLM separately in 5 separate tests (Tests 1 - Tests 5). Each test was either reset or started in a new thread and prompts were initialized and the temperature is 1.0 for all LLMs for all questions. In each test, the LLM was prompted with 7–12 questions one after another (14 questions for pediatric conjunctival diseases, 7 questions for ocular manifestations of systemic diseases in children, 9 questions for pediatric retinal disorders, and 10 questions for the rest) until the test was completed, with the LLM being instructed to provide the correct answers. If the LLM was unable to process all

10 questions at once, it was adjusted to handle batches of five questions. If the LLM did not indicate all the answers, the next test included the unanswered questions along with the entire next batch of test questions. In each instance of testing, the wording of global prompts and instructional prompts is varied to account for response-noise caused by prompt-noise. The initialization prompts and instructional prompts are outlined in Table 1.

| Test | Initialization prompt | Instructions prompt |
|---|---|---|
| Test 1 | I am a pediatric ophthalmologist. My research team aims to investigate the responses provided by ChatGPT on topics related to pediatric ophthalmology. I will now proceed with further inquiries about pediatric ophthalmology. | Please give only the question label and the letter for the correct answer. |
| Test 2 | I will present you with some multiple-choice questions. | Only respond with the correct letter choice. |
| Test 3 | We would like to assess your understanding of pediatric ophthalmology. Therefore, we have created a set of questions to inquire about your knowledge. | Only give the correct answer in your response. Do not explain your answers. |
| Test 4 | Please answer the following practice questions as if you were a doctor preparing for a licensure examination. | For each multiple-choice question, provide the correct answer without any justification. |
| Test 5 | I would like to assess your knowledge of pediatric ophthalmology by presenting a series of multiple-choice questions. | In your response, only report the question label and the corresponding answer. |

Table1. The initialization prompts and instructional prompts examples.

The test scores and distributions of the LLM were compared to the scores of three groups: undergraduate students, master's degree students, and attending physicians. Eight undergraduate students majoring in clinical medicine were randomly selected from the Medical School of Nanchang University to form the undergraduate group. Five master's students specializing in clinical ophthalmology from the First Clinical Medical College of Nanchang University and two pediatric ophthalmologists from the same institution were selected for the master's degree student group and attending physician group, respectively. Each candidate was given three hours to participate in a closed-book exam. To compare the scores of the LLM and human participants, the average scores, correlation of correct answers, and confidence in the answers were evaluated.

To quantify the overall consistency of the scoring, we calculated the standard deviations and average correlations. The average correlation was defined as the average of the upper limits of the Pearson correlation matrices for the experiments. The average correlation represented the degree of consistency in the correct scores for the experiments, where 1 indicated identical distributions, 0 corresponded to a completely random distribution, and −1 indicated a completely inverse correlation in the distribution.To quantify the stability of different LLMs in answering questions, we calculated the average correlation between the answers provided by each LLM during testing and the correct answers. Additionally, in order to gain a clearer understanding of the disparities among LLMs in pediatric ophthalmology testing, we compiled and computed their scores separately, as well as the mean correlation and respective variances.

To quantify the confidence level of answers provided by the LLMs and the three different populations, we calculated the number of correct answers for each question across all tests. For instance, if each LLM correctly answered the same question five times, the percentage of questions with all five answers being correct would increase by 1/100% (as there were 100 questions in total). Additionally, the test results were compared to the expected distribution that occurs when candidates make random guesses. When guessing randomly, the expected number of correct answers in five trials averaged to approximately $0.25 \times (5 \times 45/100) + 0.2 \times (5 \times 55/100) = 1.1125$ (45/10 questions have 4 choices, 55/100 have 5 choices), given that multiple-choice questions have five or four options. Using this value, the occurrence of correct answers for each question was estimated based on the resulting Poisson distribution.

Finally, the scores obtained from the cumulative calculations of ChatGPT (GPT-3.5 and GPT-4) and PaLM2 were compared to the scores of the human groups.

**4. Results**

4.1 The comparison between LLM scores and human scores

The comparison between LLM scores and human scores is shown in Figure 1 and Figure 2. The LLM average test score represented the average score from five separate tests, while the average scores for human groups, such as undergraduate students, graduate students, and attending physicians, were the average scores of individuals within each group. In the raw scores obtained from LLM testing, there were variations in the uncertainty of the overall score and the correctness of each answered question. It can be observed that GPT-4 covered a wider range of correct answers in the test items (Figure 1). Overall, the performance of each LLM was generally better than that of the undergraduate student group. The graduate students outperformed ChatGPT-3.5 (GPT-3.5) and PaLM2 in specialized tests such as pediatric conjunctival diseases, pediatric lacrimal system diseases, pediatric amblyopia, pediatric strabismus, and pediatric ocular tumors, and they also performed better overall than the two LLMs. The attending physicians scored higher than ChatGPT-4 (GPT-4), and all LLMs performed poorly in the specialized test for pediatric strabismus (Figures 2 and 3). It is worth mentioning that LLM may not have answered this question correctly due to the ambiguity of the correct answer or the question itself lacking significant meaning that only humans can comprehend. Interestingly, this could be a method of identifying flawed questions by utilizing a large number of LLMs to select questions that have never been answered correctly in the question bank.

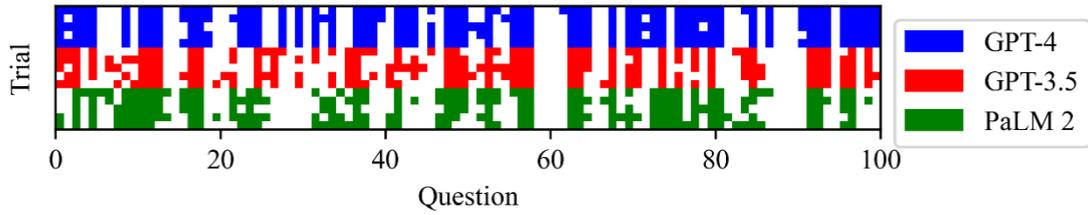

Figure1. The original annotations for each test where rows represent individual tests and columns represent test questions. Darker squares denote the correct answer.

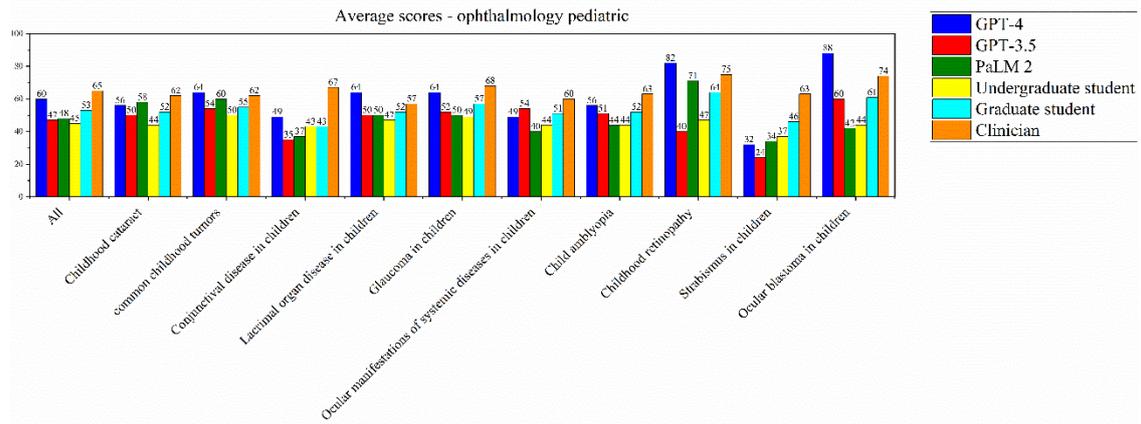

Figure 2. Mean test scores for each LLM as well as humans by category.

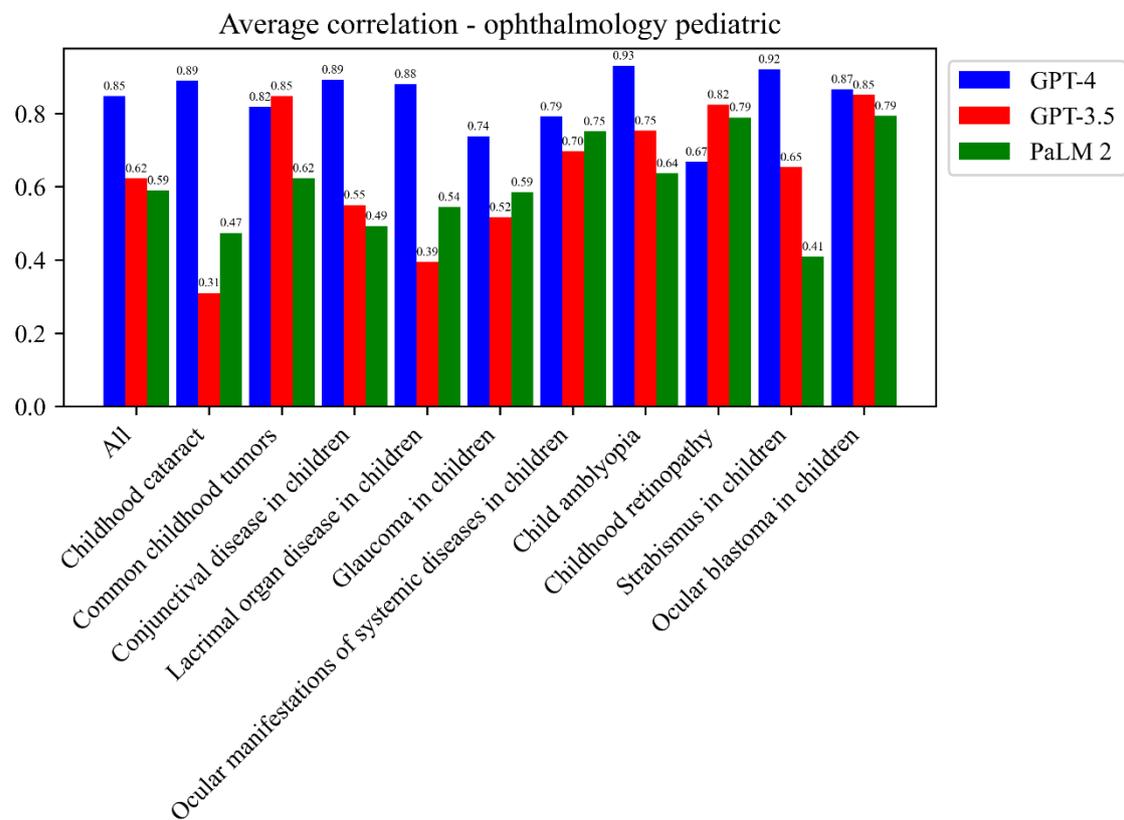

Figure 3. Differences in scores among different LLMs as well as humans.

4.2 The comparison of LLM answer stability

From Figures 4 and 5, it can be observed that ChatGPT-4 exhibited a high average correlation in the tests, consistently surpassing 0.8. It showed significant differences compared to ChatGPT-3.5 and PaLM2, with ChatGPT-4 demonstrating the least variation in the average correlation across specialized multiple-choice questions. ChatGPT-3.5 had a higher average correlation than PaLM2, but it did not exhibit significant differences. Additionally, PaLM2 showed slightly higher stability than ChatGPT-3.5 across specialized multiple-choice questions.

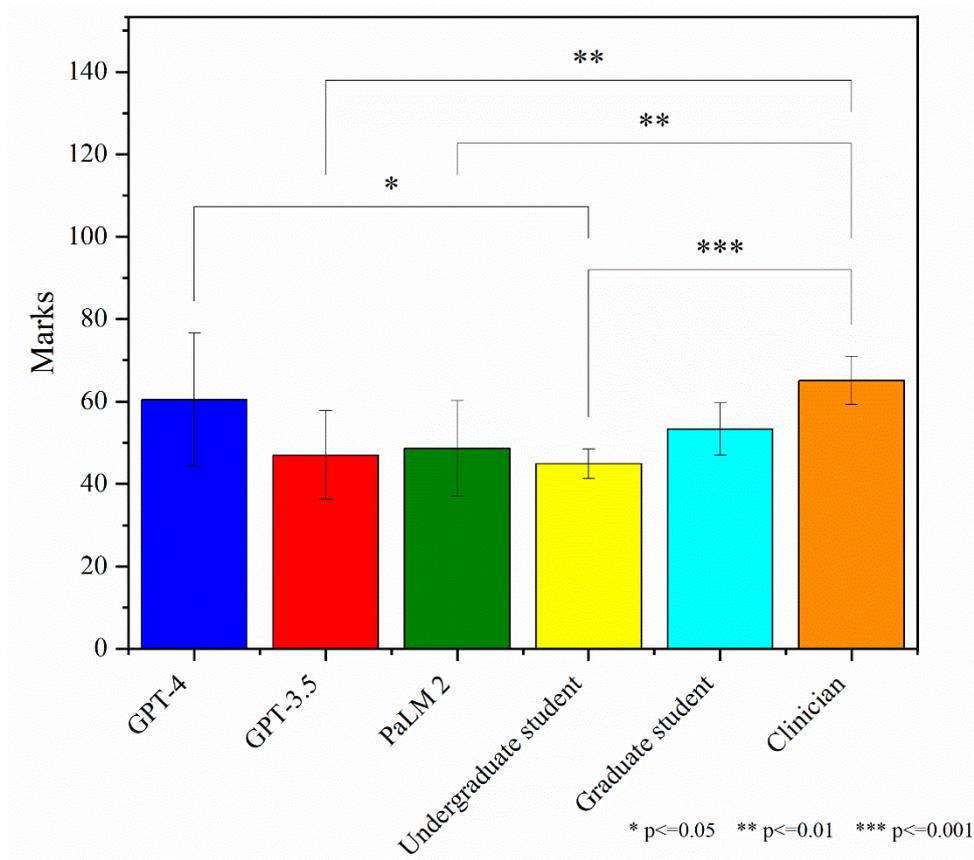

Figure 4. Mean correlation for each LLM by category.

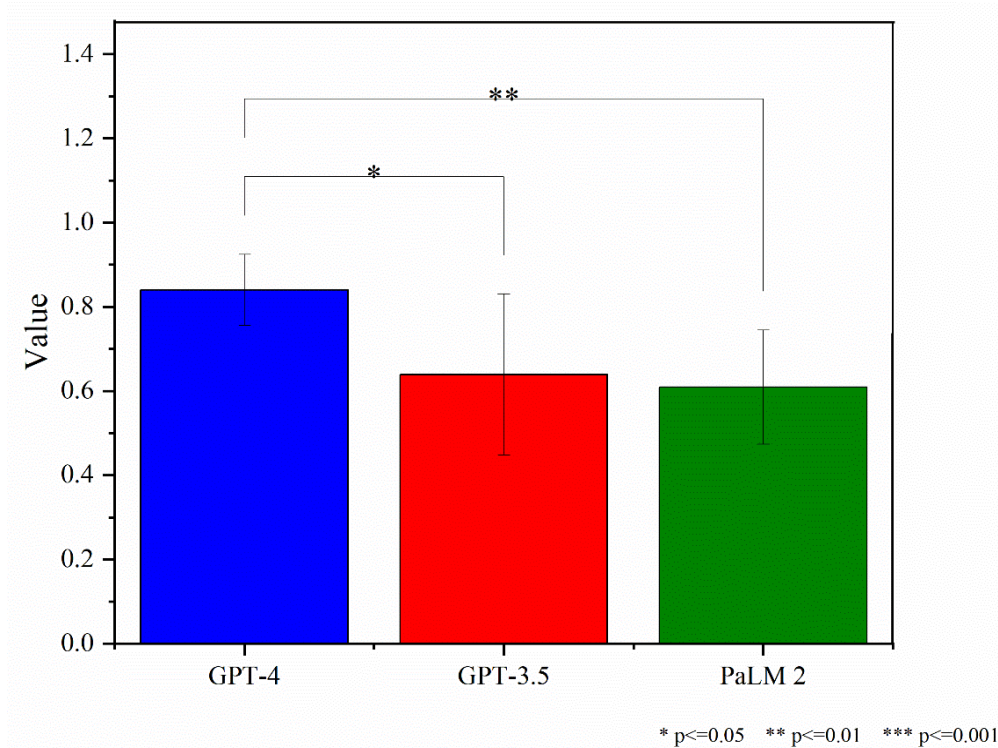

Figure 5. Differences in correlation among different LLMs.

4.3 The comparison of LLM answer confidence

From Figure 6, it can be observed that ChatGPT (ChatGPT-3.5 and ChatGPT-4) are not akin to random guessing.ChatGPT-3.5 either exhibited a high level of confidence in the answers, with 24% of the answers being correct in each test, or it showed a tendency to be confused about whether the correct answer could be determined, resulting in 34% of the answers being incorrect. Although the results indicate that LLMs show less than ideal performance, either confidently correct or confidently incorrect, our research suggests that ChatGPT-4 performs better in determining the possibility of obtaining a definitive answer, with 45% of the questions answered correctly and 37% answered incorrectly in each test. PaLM2 performed at a lower level compared to ChatGPT and exhibited a stronger tendency for confusion.

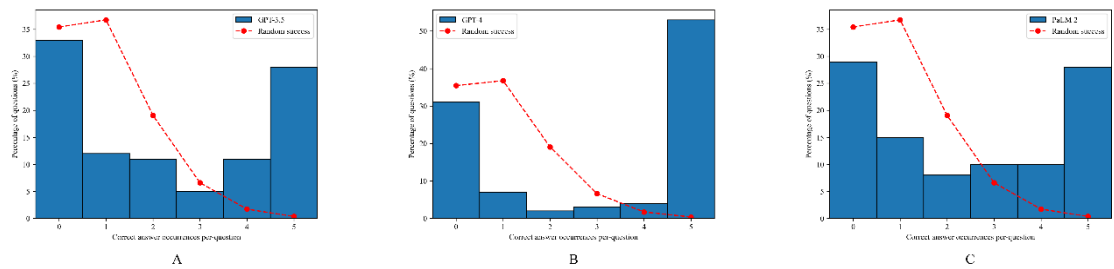

Figure 6. Confidence in answers. Frequency of correct answers in each question for both LLMs and humans groups. The red dashed line represents the expected distribution of answers based on randomly selecting according to the Poisson distribution.

## 5. Discussion

The results indicated that undergraduate students performed at lower levels than all LLMs in these highly specialized tests. Graduate students performed better than ChatGPT-3.5 but worse than ChatGPT-4. ChatGPT-4 demonstrated excellent accuracy in answering questions related to pediatric retinal lesions and pediatric ocular tumors, but performed poorly in pediatric strabismus, even scoring lower than PaLM2. Overall, its performance was comparable to that of clinical physicians. In the specialized test for pediatric strabismus, it can be observed that there was the highest proportion of diagnostic questions, accounting for 44% of the questions, compared to other tests. Particularly, in the two specialties where ChatGPT-4 performed well; i.e., pediatric retinal lesions and pediatric ocular tumors, there were either no diagnostic questions or they accounted for less than 15%. Diagnostic questions require the integration and flexible application of knowledge from various aspects, while also considering the confounding factors of false positives in pediatric diseases, which can pose challenges for LLMs and result in a decrease in their ability to answer questions. As an LLM that assists in highly specialized subject knowledge, ChatGPT-4 demonstrated superior performance and stability in terms of answer correlations and answer confidence compared to the other two LLMs. ChatGPT-3 and PaLM2 have their respective advantages in specific specialized tests, such as their significantly higher number of correct answers in pediatric ocular tumors and pediatric retinal lesions, respectively. The stability of their answers also varied across different specialized tests (Figure 4;

Figure 5), with ChatGPT-3.5 slightly outperforming PaLM2 in answer confidence (Figure 6). For assistance, ChatGPT-4 is recommended as the first choice, followed by analyzing the performance and stability of ChatGPT-3.5 and PaLM2 in specific domains before selecting the most suitable LLM. However, stability does not fully represent superior performance, and the best results would be achieved by using multiple high-performance LLMs and voting.[38] In this case, stability may be lower, but the scores will be higher. Our study only aimed to evaluate various LLMs to select the most appropriate ones for pediatric ophthalmology.

Although ChatGPT (ChatGPT-4) performed exceptionally well overall and its performance can be comparable to attending physicians in terms of answering questions (Figure 2 and 3), it does not imply that LLMs can completely replace pediatric ophthalmologists. Even with the same professional background, attending physicians have significant variations in individual abilities and knowledge, and they have more in-depth research in specific areas. Additionally, as shown in the distribution of scores in the specialized tests in Figure 6, ChatGPT exhibited absolute confidence when answering questions correctly and confusion when answering incorrectly. On the contrary, attending physicians, in highly specialized topics, know when and how to make educated guesses, even if they are uncertain about the correct answer, and their reasoning is less likely to be erroneous. However, in specific highly specialized fields like pediatric ophthalmology, ChatGPT-4 demonstrates outstanding performance, and with continuous improvement, this study suggests that ChatGPT-4 has great potential as a guide for medical students and as an assistant to attending physicians. making it a highly worthwhile subject for further research.

The present study also raises a consideration that a LLM may not fully represent the meticulous clinical work of pediatric ophthalmologists, as answering questions does not equate to the complexity encountered in daily clinical practice, leading to potential deviations in performance. Thus, in the field of pediatric ophthalmology, although a LLM, especially ChatGPT-4, holds great prospects for application, there are still numerous highly specialized areas within the medical field that may require reassessment. With the continuous advancement of LLMs, humanity will be able to devote more effort to refining cutting-edge domains rather than investing substantial resources in acquiring basic knowledge or engaging in repetitive mechanical tasks. Perhaps LLM should place greater emphasis on specialized knowledge.

5.1 The application of LLMs in pediatric ophthalmology

This study represents a continuation of applying state-of-the-art NLP methods to research in pediatric ophthalmology and related fields.[39-41] Currently, ChatGPT has been tested in the field of ophthalmology.[42] Weihao Gao et al. utilized retinal images as a starting point for disease assessment and diagnosis, achieving diagnoses of common ophthalmic diseases and lesion segmentation. Subsequently, they established a new multimodal teaching approach in ophthalmology and interacted with disease-related knowledge data to collect available real-world medical treatment plans for learning, resulting in the development of a specialized LLM for ophthalmology, namely OphGLM, which has demonstrated remarkable functionality in subsequent experiments.[43] Children, being a vulnerable and extensive population, require timely and accurate diagnosis and treatment, making online diagnosis and treatment a favorable solution. However, doctors may not always be able to meet the demand for rapid responses. Therefore, utilizing an LLM for online consultations is a promising approach, as research has shown no significant differences in correctness and safety compared to answers provided by humans.[44] The diagnosis and treatment of rare eye diseases in children worldwide is a valuable topic.[45-47] However, frequent misdiagnosis or slow improvement in treatment rates often occur due to difficulties in diagnosing disease manifestations. Liao et al.[48] proposed a language architecture that guides the attention of the Transformer language architecture to more important decision markers, enabling training of language models with fewer annotated samples. This approach may become a solution for training language models with small sample sizes in the future. Additionally, the exciting aspect is that ChatGPT, as an effective method for text data augmentation, has demonstrated superior performance in terms of testing accuracy and enhancing sample distribution compared to current state-of-the-art text data augmentation methods.[49] It also holds potential for addressing the challenges of small sample learning.

Furthermore, LLMs are also utilized in applications such as constructing network graphs and data recognition. For instance, it can be used to build domain-specific knowledge graphs without the need for manual annotations from clinical doctors or other domain experts, saving time and reducing effort.[50] Due to the challenges of data anonymization in cross-institutional clinical collaboration and academic research, there is a strong demand for clinical note recognition. ChatGPT-4 has demonstrated excellent performance in clinical note recognition, with an accuracy rate of up to 99%, surpassing other language models.[13]

5.2 The application of LLMs in medical education

Based on the above discussion, it is worth considering whether an LLM, specifically ChatGPT-4, can serve as a guiding teacher in medical education, as it demonstrates comparable performance to attending physicians in terms of question accuracy, stability, and relevance. From the experimental data, ChatGPT-4 can indeed act as a guiding teacher for beginners, but it is important to note its limitations. It lacks the ability to integrate external information, understand sensory and non-verbal cues, and cultivate harmonious interpersonal and doctor-patient relationships.[51] Additionally, in tasks such as paper writing and assignment guidance, it is challenging to ensure the quality of independent learning for medical students, as they may overly rely on an LLM instead of engaging in critical thinking, which can significantly compromise learning outcomes.[52, 53] Therefore, countermeasures need to be implemented.[54] Furthermore, patient privacy and ethical integrity are important considerations, and precautions must be taken during training to avoid data breaches and guide against negative responses.[40] Currently, the performance of LLM's in specialized medical fields, such as ophthalmology, is not particularly outstanding.[55, 56] However, we believe that with the development of LLMs, it will become a valuable tool in higher education and an indispensable learning assistant in the future[57].

5.3 Multimodal models in pediatric ophthalmology

Multimodal models are the future of language models[1] and hold exciting potential in medical diagnostics. Early research on multimodal data in LLMs includes ChatCAD,[58] which is a framework that integrates images, text, and computer-aided diagnosis. It supports various diagnostic networks, such as automated lesion image segmentation and report generation[59]. In this framework, ChatGPT greatly enhances the output of these networks.

ChatGPT-4 supports multimodal inputs, including images, which further unleashes the potential of LLM in pediatric ophthalmology. The success[15, 17, 60-64] of foundational models extend beyond NLP[14, 57, 65-68]. In future research[69], it is necessary to integrate text, images, drug dosages, and other datasets into the diagnostic and treatment processes. We believe that with advancements in neuroscience, multimodal models have shown strong similarities to the human brain,[1] and they will contribute to the progress of NLP[70].

6. Conclusion

This study demonstrated that LLMs, especially ChatGPT (ChatGPT-4), excel at answering highly specialized questions in pediatric ophthalmology, showcasing impressive capabilities in testing. While ChatGPT-4 performed at a comparable level to attending physicians, it exhibited exceptional accuracy and stability compared to undergraduate students, indicating promising prospects for its use in undergraduate medical education and knowledge sharing. With the limited availability of clinical doctors, the assistance of ChatGPT-4 can further enhance diagnostic accuracy and reduce the efficiency of misdiagnosis. In economically underdeveloped and remote areas with limited access to highly skilled doctors, the use of ChatGPT-4 can greatly support the personal development of local doctors and improve the well-being of the local population, thereby reducing the healthcare burden on major cities. In conclusion, the research results demonstrated the application potential and promising future of LLMs in specialized fields.

# Appendix

# 100 Children's Ophthalmology Multiple Choice Questions

# Correct answer: 1

# Incorrect answer: 0

Childhood cataract

1.The most common type of cataract in childhood, caused by A malfunction in the metabolism of the lens during a certain period of the embryo (see figure), is most likely to be what type of cataract ()   A. Congenital circumnuclear cataract   1

1.The most common type of cataract in childhood, caused by A malfunction in the metabolism of the lens during a certain period of the embryo (see figure), is most likely to be what type of cataract ()   B. Congenital anterior polar cataract   0

1.The most common type of cataract in childhood, caused by A malfunction in the metabolism of the lens during a certain period of the embryo (see figure), is most likely to be what type of cataract ()   C. Congenital posterior polar cataract   0

1.The most common type of cataract in childhood, caused by A malfunction in the metabolism of the lens during a certain period of the embryo (see figure), is most likely to be what type of cataract ()   D. Congenital punctate cataract     0

1.The most common type of cataract in childhood, caused by A malfunction in the metabolism of the lens during a certain period of the embryo (see figure), is most likely to be what type of cataract ()   E. None of the above 0

2.Patient, female, 5 years old, found to have an external right eye oblique for 6 months. Check the lens mix Turbidity, dilated pupil examination no abnormal fundus, the general condition is OK. History family There is no such patient, the mother has no history of cold and rubella infection during pregnancy, but often Hand and foot convulsions, when the examination had low blood calcium and high blood phosphorus.The most likely diagnosis is ()  A. Traumatic cataract 0

2.Patient, female, 5 years old, found to have an external right eye oblique for 6 months. Check the lens mix Turbidity, dilated pupil examination no abnormal fundus, the general condition is OK. History family There is no such patient, the mother has no history of cold and rubella infection during pregnancy, but often Hand and foot convulsions, when the examination had low blood calcium and high blood phosphorus.The most likely diagnosis is ()  B. Congenital cataract     1

2.Patient, female, 5 years old, found to have an external right eye oblique for 6 months. Check the lens mix Turbidity, dilated pupil examination no abnormal fundus, the general condition is OK. History family There is no such patient, the mother has no history of cold and rubella infection during pregnancy, but often Hand and foot convulsions, when the examination had low blood calcium and high blood phosphorus.The most likely diagnosis is ()  C. Metabolic cataract 0

2.Patient, female, 5 years old, found to have an external right eye oblique for 6 months. Check the lens mix Turbidity, dilated pupil examination no abnormal fundus, the general condition is OK. History family There is no such patient, the mother has no history of cold and rubella infection during pregnancy, but often Hand and foot convulsions, when the examination had low blood calcium and high blood phosphorus.The most likely diagnosis is () D. Congenital rubella syndrome     0

2.Patient, female, 5 years old, found to have an external right eye oblique for 6 months. Check the lens mix Turbidity, dilated pupil examination no abnormal fundus, the general condition is OK. History family There is no such patient, the mother has no history of cold and rubella infection during pregnancy, but often Hand and foot convulsions, when the examination had low blood calcium and high blood phosphorus.The most likely diagnosis is () E. Retinoblastoma     0

3.The Treatment method of Congenital cataract: ()   A. At present, the child is too small to be treated    0

3.The Treatment method of Congenital cataract: ()   B. The right eye was obliqued and no treatment was necessary    0

3.The Treatment method of Congenital cataract: ()   C. The cloudy lens is surgically removed and a suitable intraocular lens is implanted 1

3.The Treatment method of Congenital cataract: ()   D. The patient is too young to be implanted with an intraocular lens, and the mixture is now surgically removed The cloudy lens was implanted with an intraocular lens in the second phase after it grew     0

3.The Treatment method of Congenital cataract: ()   E. Start with lazy eye training for the right eye 0

4.Congenital cataract's Prognosis of children () A. There is strabismus, and the vision will not improve after surgical treatment     0

4.Congenital cataract's Prognosis of children ()B. Surgery will remove the cloudy lens, and the vision will naturally improve    0

4.Congenital cataract's Prognosis of children () C. After surgery, the cloudy lens is removed and then treated with glasses, the vision will be improved    0

4.Congenital cataract's Prognosis of children () D. Surgically, the cloudy lens is removed and implanted into an intraocular lens enhance 0

4.Congenital cataract's Prognosis of children () E. The cloudy lens is surgically removed and then implanted into the intraocular lens After a period of rigorous amblyopia training your vision may improve considerably    1

5.What eye lesions in newborns caused by measles infection in mothers during the first 3 months of pregnancy () A. Congenital cataract    1

5.What eye lesions in newborns caused by measles infection in mothers during the first 3 months of pregnancy () B. Infantile glaucoma 0

5.What eye lesions in newborns caused by measles infection in mothers during the first 3 months of pregnancy () C. erson syndrome    0

5.What eye lesions in newborns caused by measles infection in mothers during the first 3 months of pregnancy () D. Valsalva syndrome     0

6.The most common cause of white pupil syndrome is() A  Coat's  disease  (retinal telangiectasia)    0

6.The most common cause of white pupil syndrome is() B retinopathy of prematurity

0

6.The most common cause of white pupil syndrome is()　C congenital cataract　1
6.The most common cause of white pupil syndrome is()　D retinoblastoma　0
6.The most common cause of white pupil syndrome is()　E retinal vasculitis　0
7.Which of the following is not an indication for intracapsular extraction ()　A traumatic cataract without rupture of capsule　0
7.Which of the following is not an indication for intracapsular extraction ()　B complicated cataract　0
7.Which of the following is not an indication for intracapsular extraction ()　C infantile cataract　1
7.Which of the following is not an indication for intracapsular extraction ()　D lens dislocation or subluxation　0
7.Which of the following is not an indication for intracapsular extraction ()　E senile cataract　0
8.The most valuable examination for congenital cataract in the left eye is()　A VEP　0
8.The most valuable examination for congenital cataract in the left eye is()　B UBM　0
8.The most valuable examination for congenital cataract in the left eye is()　C eye B ultrasound　1
8.The most valuable examination for congenital cataract in the left eye is()　D orbital CT　0
8.The most valuable examination for congenital cataract in the left eye is()　E ERG　0
9.If the child has cataracts in the left eye, which of the following arguments is more appropriate()A cataract removal in the left eye as soon as possible, implantation of intraocular lens in adults　0
9.If the child has cataracts in the left eye, which of the following arguments is more appropriate()B surgery is not considered for the time being, cataract removal combined with intraocular lens implantation after 7 years of age　0
9.If the child has cataracts in the left eye, which of the following arguments is more appropriate()C lens removal in the left eye as soon as possible　0
9.If the child has cataracts in the left eye, which of the following arguments is more appropriate()D lens removal in the left eye as soon as possible, and active treatment of amblyopia after surgery. After 3 years of age, human crystal　1
9.If the child has cataracts in the left eye, which of the following arguments is more appropriate()E can be considered for implantation and ECCE+IOL when adults grow up　0
10.Exophthalmos, blepharospasm, refractive errors and hypocalcic cataracts in infants and young children ()A. Vitamin A deficiency　0
10.Exophthalmos, blepharospasm, refractive errors and hypocalcic cataracts in infants and young children ()B. Vitamin B deficiency　0
10.Exophthalmos, blepharospasm, refractive errors and hypocalcic cataracts in infants and young children ()C. Vitamin B deficiency　0

10.Exophthalmos, blepharospasm, refractive errors and hypocalcic cataracts in infants and young children () D. Vitamin C deficiency   0

10.Exophthalmos, blepharospasm, refractive errors and hypocalcic cataracts in infants and young children () E. Vitamin D deficiency   1

Common childhood tumors

11.The most common primary orbital malignancies in childhood are() A.     Optic neuromeningioma     0

11.The most common primary orbital malignancies in childhood are() B. Optic glioma     0

11.The most common primary orbital malignancies in childhood are() C.     Orbital dermoid cyst 0

11.The most common primary orbital malignancies in childhood are() D. rhabdomyosarcoma   1

11.The most common primary orbital malignancies in childhood are() E.     cavernous hemangioma 0

12.The most common tumors of the optic nerve in childhood are()     A.     Optic neuromeningioma     0

12.The most common tumors of the optic nerve in childhood are()     B. Optic glioma     1

12.The most common tumors of the optic nerve in childhood are()     C.     Orbital dermoid cyst     0

12.The most common tumors of the optic nerve in childhood are()     D. rhabdomyosarcoma   0

12.The most common tumors of the optic nerve in childhood are()     E.     cavernous hemangioma 0

13.The eyelid skin bumps in infants and young children are bright red or dark red patches, flat and raised, should be ()    A. dermoid cyst  0

13.The eyelid skin bumps in infants and young children are bright red or dark red patches, flat and raised, should be ()    B. capillary hemangioma  1

13.The eyelid skin bumps in infants and young children are bright red or dark red patches, flat and raised, should be ()    C. Cavernous hemangioma     0

13.The eyelid skin bumps in infants and young children are bright red or dark red patches, flat and raised, should be ()    D. plasmacytoma 0

13.The eyelid skin bumps in infants and young children are bright red or dark red patches, flat and raised, should be ()    E. xanthoma 0

14.A male, 8 years old, presented with blurred vision in the left eye for 1 week with pain from eye movements and a prior history of colds. Check pupil light reflex slightly dull, fundus examination ()     A. Triple mirror examination   0

14.A male, 8 years old, presented with blurred vision in the left eye for 1 week with pain from eye movements and a prior history of colds. Check pupil light reflex slightly dull, fundus examination ()    B. optometry      0

14.A male, 8 years old, presented with blurred vision in the left eye for 1 week with pain from eye movements and a prior history of colds. Check pupil light reflex slightly dull, fundus examination ()    C. ultrasound examination     0

14. A male, 8 years old, presented with blurred vision in the left eye for 1 week with pain from eye movements and a prior history of colds. Check pupil light reflex slightly dull, fundus examination ()    D. head CT examination   0

14. A male, 8 years old, presented with blurred vision in the left eye for 1 week with pain from eye movements and a prior history of colds. Check pupil light reflex slightly dull, fundus examination ()    E. perimetry   1

15. A patient, 8 years old, presented with ophthalmic atrophy in the fundus and painless progressive loss of visual acuity due to protrusion in both eyes. The patient was most likely to have ()   A. Glioma   1

15. A patient, 8 years old, presented with ophthalmic atrophy in the fundus and painless progressive loss of visual acuity due to protrusion in both eyes. The patient was most likely to have ()   B. Optic nerve meningioma   0

15. A patient, 8 years old, presented with ophthalmic atrophy in the fundus and painless progressive loss of visual acuity due to protrusion in both eyes. The patient was most likely to have ()   C. Optic papillary hemangioma   0

15. A patient, 8 years old, presented with ophthalmic atrophy in the fundus and painless progressive loss of visual acuity due to protrusion in both eyes. The patient was most likely to have ()   D. Neurofibroma   0

15. A patient, 8 years old, presented with ophthalmic atrophy in the fundus and painless progressive loss of visual acuity due to protrusion in both eyes. The patient was most likely to have ()   E. None of the above   0

16. Green tumors occur more often in ()    A. Baby   1
16. Green tumors occur more often in ()    B. Teen   0
16. Green tumors occur more often in ()    C. Adult   0
16. Green tumors occur more often in ()    D. Old man   0
16. Green tumors occur more often in ()    E. Young women   0

17. The most common benign eyelid tumors in infants and young children are()   A   molluscum contagiosus   0
17. The most common benign eyelid tumors in infants and young children are()   B   basal cell papilloma   0
17. The most common benign eyelid tumors in infants and young children are()   C   xanthoma   0
17. The most common benign eyelid tumors in infants and young children are()   D   keratoacanthoma   0
17. The most common benign eyelid tumors in infants and young children are()   E   capillary hemangioma   1

18. Most common intraorbital malignancies in childhood()   A   rhabdomyosarcoma   1
18. Most common intraorbital malignancies in childhood()   B   leiomyosarcoma   0
18. Most common intraorbital malignancies in childhood()   C   chondrosarcoma   0
18. Most common intraorbital malignancies in childhood()   D   Malignant meningioma   0
18. Most common intraorbital malignancies in childhood()   E   fibrosarcoma   0

19. Twenty-two. A patient, 8 years old, presented to the doctor due to binocular

protrusion, progressive painless vision loss, optic nerve atrophy in the fundus, and orbital CT images. The patient was most likely to suffer from()　A. Glioma　1

19.Twenty-two. A patient, 8 years old, presented to the doctor due to binocular protrusion, progressive painless vision loss, optic nerve atrophy in the fundus, and orbital CT images. The patient was most likely to suffer from()　B.　Optic　nerve meningioma　0

19.Twenty-two. A patient, 8 years old, presented to the doctor due to binocular protrusion, progressive painless vision loss, optic nerve atrophy in the fundus, and orbital CT images. The patient was most likely to suffer from()　C. Hemangioma of the optic papilla　0

19.Twenty-two. A patient, 8 years old, presented to the doctor due to binocular protrusion, progressive painless vision loss, optic nerve atrophy in the fundus, and orbital CT images. The patient was most likely to suffer from()　D.　Neurofibroma　0

19.Twenty-two. A patient, 8 years old, presented to the doctor due to binocular protrusion, progressive painless vision loss, optic nerve atrophy in the fundus, and orbital CT images. The patient was most likely to suffer from()　E. None of the above　0

20.The following statement is false()　A. Glioma of the eye is more common in children　0

20.The following statement is false()　B. The edema of optic disc is mostly unilateral　1

20.The following statement is false()　C. The visual field in the early stage of acute disc edema expands the physiological blind spot　0

20.The following statement is false()　D. Optic nerve meningioma is more common in adults　0

20.The following statement is false()　E. Increased intracranial pressure is the most common cause of optic disc edema　0

Conjunctival disease in children

21. Most of the pathogens causing neonatal dacryocystitis are ()　A.　Staphylococcus aureus　0

21. Most of the pathogens causing neonatal dacryocystitis are ()　B.　Streptococcus hemolyticus　0

21. Most of the pathogens causing neonatal dacryocystitis are ()　C. Candida albicans　0

21. Most of the pathogens causing neonatal dacryocystitis are ()　D.　Haemophilus influenzae　1

21. Most of the pathogens causing neonatal dacryocystitis are ()　E. All of the above　0

22.What is known as "pus leaking eyes" ()　A. Adult gonococcal conjunctivitis　0

22.What is known as "pus leaking eyes" ()　B. Neonatal meningococcal conjunctivitis　0

22.What is known as "pus leaking eyes" ()　C. Meningococcal conjunctivitis in adults　0

22.What is known as "pus leaking eyes" ()  D. Neonatal gonococcal conjunctivitis  1
22.What is known as "pus leaking eyes" ()  E. Acute conjunctivitis by type     0
23. The incubation period of neonatal gonococcal conjunctivitis is ()  A. 2 to 5 days   1
23. The incubation period of neonatal gonococcal conjunctivitis is ()  B. within 24 hours   0
23. The incubation period of neonatal gonococcal conjunctivitis is ()  C. Minutes   0
23. The incubation period of neonatal gonococcal conjunctivitis is ()  D. 6 to 15 days   0
23. The incubation period of neonatal gonococcal conjunctivitis is ()  E. 16 to 30 days   0
24. Neisser meningococcal conjunctivitis ()A. More women than men     0
24. Neisser meningococcal conjunctivitis ()B. Adults are more common than children   0
24. Neisser meningococcal conjunctivitis ()C. Men are more common than women 0
24. Neisser meningococcal conjunctivitis ()D. Children are more common than adults   1
24. Neisser meningococcal conjunctivitis ()E. The incidence is equal in children and adults   0
25. The following incorrect description of the clinical features of optic neuritis (see figure) is ()  A. Visual impairment is most severe within 1 week of onset  0
25. The following incorrect description of the clinical features of optic neuritis (see figure) is ()  B. The incidence of binocular involvement is higher in adults than in children 1
25. The following incorrect description of the clinical features of optic neuritis (see figure) is ()  C. Pain with eye movement   0
25. The following incorrect description of the clinical features of optic neuritis (see figure) is ()  D. Have central scotoma or reduced visual field centripetal   0
25. The following incorrect description of the clinical features of optic neuritis (see figure) is ()  E. VEP is characterized by prolonged latency and decreased amplitude of P wave (P1 wave)    0
26. A male, 7 years old, had blurred vision in the left eye for 1 week with pain when moving the eye and a prior history of cold. The pupil was slightly slow in light reflex,The diagnosis is()  A. Retrobulbar neuritis   1
26. A male, 7 years old, had blurred vision in the left eye for 1 week with pain when moving the eye and a prior history of cold. The pupil was slightly slow in light reflex,The diagnosis is()  B. Ischemic optic neuropathy  0
26. A male, 7 years old, had blurred vision in the left eye for 1 week with pain when moving the eye and a prior history of cold. The pupil was slightly slow in light reflex,The diagnosis is()  C. Scleritis   0
26. A male, 7 years old, had blurred vision in the left eye for 1 week with pain when moving the eye and a prior history of cold. The pupil was slightly slow in light reflex,The diagnosis is()  D. Refractive error    0
26. A male, 7 years old, had blurred vision in the left eye for 1 week with pain when

moving the eye and a prior history of cold. The pupil was slightly slow in light reflex,The diagnosis is()   E. Iridocyclitis    0
27.The patient, female, 6 years old, had acute parotitis 1 week ago. The fundus examination was shown in the figure. The wrong clinical characteristics were ()     A.VEP is characterized by prolonged latency and decreased amplitude of P wave (P1 wave)    0
27.The patient, female, 6 years old, had acute parotitis 1 week ago. The fundus examination was shown in the figure. The wrong clinical characteristics were ()     B. Pain with eye movement   0
27.The patient, female, 6 years old, had acute parotitis 1 week ago. The fundus examination was shown in the figure. The wrong clinical characteristics were ()     C. The incidence of binocular involvement is higher in adults than in children. Have central scotoma or reduced visual field centripetal    1
27.The patient, female, 6 years old, had acute parotitis 1 week ago. The fundus examination was shown in the figure. The wrong clinical characteristics were ()     D. Visual impairment is most severe at 1 week of onset       0
28.The pathogenic bacteria of newborn pyorrhea is()     A Herpes simplex virus    0
28.The pathogenic bacteria of newborn pyorrhea is()     B filamentous fungi   0
28.The pathogenic bacteria of newborn pyorrhea is()     C gonococcus     1
28.The pathogenic bacteria of newborn pyorrhea is()     D Pyocyanobacter     0
28.The pathogenic bacteria of newborn pyorrhea is()     E Acanthamoeba      0
29.The onset time of neonatal gonococcal conjunctivitis()     A 2 to 3 days after the birth    1
29.The onset time of neonatal gonococcal conjunctivitis()     B within 1 week after the birth    0
29.The onset time of neonatal gonococcal conjunctivitis()     C within 24 hours after the birth      0
29.The onset time of neonatal gonococcal conjunctivitis()     D within 1 month after the birth 0
29.The onset time of neonatal gonococcal conjunctivitis()     E within 5 days   0
30.For pregnant women, nursing women, and children under 1 year old after trachoma should not be used()   A spiramycin       0
30.For pregnant women, nursing women, and children under 1 year old after trachoma should not be used()   B doxycycline     0
30.For pregnant women, nursing women, and children under 1 year old after trachoma should not be used()   C Avantamycin   0
30.For pregnant women, nursing women, and children under 1 year old after trachoma should not be used()   D erythromycin   0
30.For pregnant women, nursing women, and children under 1 year old after trachoma should not be used()   E Tetracycline     1
31.The newborn should be routinely applied after birth() A   0.5%   Tetracycline   Eye cream   1
31.The newborn should be routinely applied after birth() B Betosol Eye Drops       0
31.The newborn should be routinely applied after birth() C Betosol Eye drops  0

31. The newborn should be routinely applied after birth( ) D EDTA Eye drops 0

32. The most common cause of unilateral exophthalmos in children is( ) A orbital cellulitis 1

32. The most common cause of unilateral exophthalmos in children is( ) B retinoblastoma 0

32. The most common cause of unilateral exophthalmos in children is( ) C rhabdomyosarcoma 0

32. The most common cause of unilateral exophthalmos in children is( ) D optic neuromeningioma 0

32. The most common cause of unilateral exophthalmos in children is( ) E optic glioma 0

33. Neisser meningococcal conjunctivitis( ) A. More women than men 0

33. Neisser meningococcal conjunctivitis( ) B. Adults are more common than children 0

33. Neisser meningococcal conjunctivitis( ) C. More men than women 0

33. Neisser meningococcal conjunctivitis( ) D. Children are more common than adults 1

33. Neisser meningococcal conjunctivitis( ) E The incidence is equal in children and adults 0

34. Pseudoconjunctiva may not be seen in( )A. Primary herpes simplex virus conjunctivitis 0

34. Pseudoconjunctiva may not be seen in( )B. beta - hemolytic streptococcal conjunctivitis 0

34. Pseudoconjunctiva may not be seen in( )C. Infantile adenovirus epidemic keratoconjunctivitis 0

34. Pseudoconjunctiva may not be seen in( )D. diphtheria conjunctivitis 1

Lacrimal organ disease in children

35. Conservative treatment has no effect. Considering lacrimal passage probing is ( ) A. Functional dacryorrhea 0

35. Conservative treatment has no effect. Considering lacrimal passage probing is ( ) B. Obstruction of the lacrimal passage in infants 1

35. Conservative treatment has no effect. Considering lacrimal passage probing is ( ) C. Chronic dacryocystitis 0

35. Conservative treatment has no effect. Considering lacrimal passage probing is ( ) D. Narrow lacrimal spots 0

35. Conservative treatment has no effect. Considering lacrimal passage probing is ( ) E. Acute dacryocystitis 0

36. The main causes of tears in infants and young children are( ) A. conjunctivitis 0

36. The main causes of tears in infants and young children are( ) B. Congenital lacrimal atresia 0

36. The main causes of tears in infants and young children are( ) C. Congenital tear duct atresia 0

36. The main causes of tears in infants and young children are( ) D. Increased tear secretion 0

36. The main causes of tears in infants and young children are ()    E.    Hypoplasia    of inferior nasolacrimal duct    1
37. The most common cause of neonatal dacryocystitis is ()    A.    Diplococcus pneumoniae    0
37. The most common cause of neonatal dacryocystitis is ()    B. Streptococcus    0
37. The most common cause of neonatal dacryocystitis is ()    C.    Haemophilus influenzae    1
37. The most common cause of neonatal dacryocystitis is ()    D.    Staphylococcus aureus    0
37. The most common cause of neonatal dacryocystitis is ()    E. E. coli    0
38. The main causes of neonatal dacryocystitis are ()    A.    Primary    infection    of    the dacryocyst    0
38. The main causes of neonatal dacryocystitis are ()    B.    Secondary    infection    of lacrimal sac caused by hypoplasia of lower segment of nasolacrimal duct    1
38. The main causes of neonatal dacryocystitis are ()    C. Congenital  canalicular  atresia    0
38. The main causes of neonatal dacryocystitis are ()    D. Infection of the birth canal    0
38. The main causes of neonatal dacryocystitis are ()    E. Congenital lacrimal atresia    0
39. Therapeutic lacrimal passage probing is mainly used for ()    A.    Lacrimal    duct obstruction    0
39. Therapeutic lacrimal passage probing is mainly used for ()    B. Lacrimal  passage obstruction in infants    1
39. Therapeutic lacrimal passage probing is mainly used for ()    C. Duct  obstruction    0
39. Therapeutic lacrimal passage probing is mainly used for ()    D.    Adult nasolacrimal duct obstruction    0
39. Therapeutic lacrimal passage probing is mainly used for ()    E.    Lacrimal obstruction    0
40. The error in the treatment of obstructed lacrimal passage in infants is ()A.  Surgical treatment as soon as possible to avoid the development of chronic dacryocystitis    1
40. The error in the treatment of obstructed lacrimal passage in infants is ()B.    Apply pressure to the lacrimal sac area    0
40. The error in the treatment of obstructed lacrimal passage in infants is ()C with your fingers. Take antibiotic eye fluid    0
40. The error in the treatment of obstructed lacrimal passage in infants is ().    Lacrimal passage probing is feasible after half a year of age.    0
40. The error in the treatment of obstructed lacrimal passage in infants is ()E    Pressure should be maintained on the lacrimal sac area for several weeks    0
41. The incorrect treatment of lacrimal passage obstruction in infants and young children is () A. Dacryocyst area massage    0
41. The incorrect treatment of lacrimal passage obstruction in infants and young children is () B. Take antibiotic eye drops    0
41. The incorrect treatment of lacrimal passage obstruction in infants and young children is () C. If conservative treatment does not work, dacryocystonasal anastomosis

may be considered after half a year    1
41.The incorrect treatment of lacrimal passage obstruction in infants and young children is () D.If conservative treatment does not work, exploratory lacrimal passage may be considered after half a year    0
42.The bacteria that cause acute dacryocystitis in infants are()    A    β    hemolytic Streptococcus    0
42.The bacteria that cause acute dacryocystitis in infants are()    B pneumococci    0
42.The bacteria that cause acute dacryocystitis in infants are()    C    Candida    albicans    0
42.The bacteria that cause acute dacryocystitis in infants are()    D    Staphylococcus aureus    0
42.The bacteria that cause acute dacryocystitis in infants are()    E    Haemophilus influenzae    1
43.The main causes of neonatal dacryocystitis are() A.    Primary    infection    of    the dacryocyst    0
43.The main causes of neonatal dacryocystitis are()  B. Secondary infection of lacrimal sac caused by hypoplasia of lower segment of nasolacrimal duct    1
43.The main causes of neonatal dacryocystitis are()  C.    Congenital    canalicular    atresia    0
43.The main causes of neonatal dacryocystitis are()  D. Infection of the birth canal   0
43.The main causes of neonatal dacryocystitis are()  E. Congenital    atresia    of    lacrimal spots    0
44.What is wrong in the treatment of obstruction of lacrimal passage in infants is() A, surgical treatment as soon as possible to avoid the development of chronic dacryocystitis    1
44.What is wrong in the treatment of obstruction of lacrimal passage in infants is() B. Press the lacrimal sac area with your fingers    0
44.What is wrong in the treatment of obstruction of lacrimal passage in infants is() C. Apply antibiotic eye fluid 0
44.What is wrong in the treatment of obstruction of lacrimal passage in infants is() D. Lacrimal passage probing is feasible after half a year    0
44.What is wrong in the treatment of obstruction of lacrimal passage in infants is() E. The lacrimal sac area should be pressed for several weeks    0
Glaucoma in children
45.The best surgical procedure for infantile glaucoma with undefined angular structure is () A. periiridectomy    0
45.The best surgical procedure for infantile glaucoma with undefined angular structure is () B. trabeculectomy    1
45.The best surgical procedure for infantile glaucoma with undefined angular structure is () C. angularotomy 0
45.The best surgical procedure for infantile glaucoma with undefined angular structure is () D. trabeculectomy    0
45.The best surgical procedure for infantile glaucoma with undefined angular structure is () E. Drainage valve implantation    0

46. Infantile glaucoma is seen in () A. 1 year of age  0
46. Infantile glaucoma is seen in () B. 3 years of age 1
46. Infantile glaucoma is seen in () C. Excludes the neonatal period  0
46. Infantile glaucoma is seen in () D. 1 to 3 years  0
46. Infantile glaucoma is seen in () E. within 6 years  0
47. The three major symptoms of non-infantile glaucoma are () A. Large eyeball, large cornea  1
47. The three major symptoms of non-infantile glaucoma are () B. Photophobia  0
47. The three major symptoms of non-infantile glaucoma are () C. Tears 0
47. The three major symptoms of non-infantile glaucoma are () D. Blepharospasm  0
48. Infantile glaucoma begins with () A. Angiotomy  1
48. Infantile glaucoma begins with () B. Trabeculectomy  0
48. Infantile glaucoma begins with () C. Nonpenetrating trabeculectomy  0
48. Infantile glaucoma begins with () D. Laser trabeculectomy  0
49. Patient, female, 2 years old, right eye photophobia, tears for 3 months. Check right corner Large membrane diameter, corneal edema, opacity, deep anterior chamber, intraocular pressure 45mmHg. The most likely diagnosis is () A. Neonatal dacryocystitis  0
49. Patient, female, 2 years old, right eye photophobia, tears for 3 months. Check right corner Large membrane diameter, corneal edema, opacity, deep anterior chamber, intraocular pressure 45mmHg. The most likely diagnosis is () B. macrocornea  0
49. Patient, female, 2 years old, right eye photophobia, tears for 3 months. Check right corner Large membrane diameter, corneal edema, opacity, deep anterior chamber, intraocular pressure 45mmHg. The most likely diagnosis is () C. Acute conjunctivitis  0
49. Patient, female, 2 years old, right eye photophobia, tears for 3 months. Check right corner Large membrane diameter, corneal edema, opacity, deep anterior chamber, intraocular pressure 45mmHg. The most likely diagnosis is () D. Hereditary corneal epithelial dystrophy  0
49. Patient, female, 2 years old, right eye photophobia, tears for 3 months. Check right corner Large membrane diameter, corneal edema, opacity, deep anterior chamber, intraocular pressure 45mmHg. The most likely diagnosis is () E. Congenital glaucoma  1
50. The preferred treatment for congenital glaucoma is () A. Drug therapy  0
50. The preferred treatment for congenital glaucoma is () B. Follow-up observation 0
50. The preferred treatment for congenital glaucoma is () C. Trabeculectomy or anthotomy  1
50. The preferred treatment for congenital glaucoma is () D. trabeculectomy  0
50. The preferred treatment for congenital glaucoma is () E. Antibiotic eye drops  0
51. The description of congenital glaucoma is as follows: A histopathological findings of anterior iris root attachment point  0
51. The description of congenital glaucoma is as follows: B glaucomatous papillary depression of optic nerve has diagnostic significance  0

51. The description of congenital glaucoma is as follows()　　C can be divided into infantile type, adolescent type and glaucoma combined with other congenital abnormalities　　0
51. The description of congenital glaucoma is as follows()　　D mainly treated with surgery　0
51. The description of congenital glaucoma is as follows()　　E is related to abnormal development of anterior atrial Angle　　0
51. The description of congenital glaucoma is as follows()　　F all correct　1
52. Normal infant cornea transverse diameter()　A, 10mm　　0
52. Normal infant cornea transverse diameter()　B, 10.5mm　1
52. Normal infant cornea transverse diameter()　C, 11mm　　0
52. Normal infant cornea transverse diameter()　D, 11.5mm　0
52. Normal infant cornea transverse diameter()　E, 12mm　　0
53. How much transverse diameter of cornea in infants is called large cornea() A, 10mm　0
53. How much transverse diameter of cornea in infants is called large cornea() B, 10.5mm 0
53. How much transverse diameter of cornea in infants is called large cornea() C, 11mm　0
53. How much transverse diameter of cornea in infants is called large cornea() D, 11.5mm 0
53. How much transverse diameter of cornea in infants is called large cornea() E, 12mm　1
54. The patient was a 2-year-old female with photophobia in her right eye and tears for 3 months. Examination: Corneal diameter of the right eye was large, corneal edema, opacity, deep anterior chamber, intraocular pressure 45mmHg. The most likely diagnosis is()　A. Neonatal dacryocystitis　　0
54. The patient was a 2-year-old female with photophobia in her right eye and tears for 3 months. Examination: Corneal diameter of the right eye was large, corneal edema, opacity, deep anterior chamber, intraocular pressure 45mmHg. The most likely diagnosis is()　B, large cornea　0
54. The patient was a 2-year-old female with photophobia in her right eye and tears for 3 months. Examination: Corneal diameter of the right eye was large, corneal edema, opacity, deep anterior chamber, intraocular pressure 45mmHg. The most likely diagnosis is()　C. Acute conjunctivitis　　0
54. The patient was a 2-year-old female with photophobia in her right eye and tears for 3 months. Examination: Corneal diameter of the right eye was large, corneal edema, opacity, deep anterior chamber, intraocular pressure 45mmHg. The most likely diagnosis is()　D. Hereditary corneal epithelial dystrophy　0
54. The patient was a 2-year-old female with photophobia in her right eye and tears for 3 months. Examination: Corneal diameter of the right eye was large, corneal edema, opacity, deep anterior chamber, intraocular pressure 45mmHg. The most likely diagnosis is()　E. Congenital glaucoma　1

Ocular manifestations of systemic diseases in children

55. Vitamin A deficiency in infants and young children with blood vitamin A levels below (E) "g/L"().   A, 100    0
55. Vitamin A deficiency in infants and young children with blood vitamin A levels below (E) "g/L"().   B. 200    0
55. Vitamin A deficiency in infants and young children with blood vitamin A levels below (E) "g/L"().   C, 300    0
55. Vitamin A deficiency in infants and young children with blood vitamin A levels below (E) "g/L"().   D, 400    0
55. Vitamin A deficiency in infants and young children with blood vitamin A levels below (E) "g/L"().   E, 500    1
56. Exophthalmos, blepharospasm, refractive errors, and hypocalcic cataracts in infants and young children()   A. Between the outer plexiform layer and the inner core layer of the retina    0
56. Exophthalmos, blepharospasm, refractive errors, and hypocalcic cataracts in infants and young children()   B, between the outer plexiform layer of the retina and the outer nuclear layer 0
56. Exophthalmos, blepharospasm, refractive errors, and hypocalcic cataracts in infants and young children()   C. The nerve fiber layer of the retina    0
56. Exophthalmos, blepharospasm, refractive errors, and hypocalcic cataracts in infants and young children()   D, between the inner plexiform layer and the inner core layer of the retina    0
56. Exophthalmos, blepharospasm, refractive errors, and hypocalcic cataracts in infants and young children()   E, between the inner plexiform layer of the retina and the ganglion cell layer    1
57. Maternal measles infection in the first 3 months of pregnancy can cause eye lesions in newborns()    A, congenital cataract diagram    1
57. Maternal measles infection in the first 3 months of pregnancy can cause eye lesions in newborns()    B. Infantile glaucoma 0
57. Maternal measles infection in the first 3 months of pregnancy can cause eye lesions in newborns()    C. Terson syndrome    0
57. Maternal measles infection in the first 3 months of pregnancy can cause eye lesions in newborns()    D. Valsalva syndrome    0
58. Treatment of retinopathy of prematurity includes()    A, laser photocoagulation 0
58. Treatment of retinopathy of prematurity includes()    B, freezing    0
58. Treatment of retinopathy of prematurity includes()    C. Vitreous surgery    0
58. Treatment of retinopathy of prematurity includes()    D. All of the above    1
59. The tests and signs that are important for the diagnosis of congenital syphilitic keratitis syndrome are()    A. Serum Cornwall reaction positive    0
59. The tests and signs that are important for the diagnosis of congenital syphilitic keratitis syndrome are()    B. The child's parents have syphilis    0
59. The tests and signs that are important for the diagnosis of congenital syphilitic keratitis syndrome are()    C. Hearing impairment    0
59. The tests and signs that are important for the diagnosis of congenital syphilitic keratitis syndrome are()    D. All the above are correct    1

60. The time to complete development of the suspensory ligament of the crystal is ( )
A The third month of the embryo    0
60. The time to complete development of the suspensory ligament of the crystal is ( )
B Embryo 5 months    0
60. The time to complete development of the suspensory ligament of the crystal is ( )
C When the fetus is born    1
60. The time to complete development of the suspensory ligament of the crystal is ( )
D None of the above    0
60. The time to complete development of the suspensory ligament of the crystal is ( )
E Embryo 4 months    0
61. The main manifestations of corneal malacia are ( ) A The child was emaciated and mentally depressed    0
61. The main manifestations of corneal malacia are ( ) B The early symptoms are night blindness    0
61. The main manifestations of corneal malacia are ( ) C Corneal epithelium is dry, dull, insensitive, epithelial exfoliation, stromal necrosis    0
61. The main manifestations of corneal malacia are ( ) D can appear Bitot spots    0
61. The main manifestations of corneal malacia are ( ) E seems to be all right    1
Child amblyopia
62. It is currently believed that when is the earliest procedure for intraocular lens implantation in children generally performed ( ) A. 3 years old    0
62. It is currently believed that when is the earliest procedure for intraocular lens implantation in children generally performed ( ) B. 6 months    0
62. It is currently believed that when is the earliest procedure for intraocular lens implantation in children generally performed ( ) C. 1 year old    0
62. It is currently believed that when is the earliest procedure for intraocular lens implantation in children generally performed ( ) D. 2 years old    1
62. It is currently believed that when is the earliest procedure for intraocular lens implantation in children generally performed ( ) E. 5 years old    0
63. Intraocular lens implantation in children is generally done at least a few years of age ( )    A. 1 year old    0
63. Intraocular lens implantation in children is generally done at least a few years of age ( )    B. 2 years old    1
63. Intraocular lens implantation in children is generally done at least a few years of age ( )    C. 3 years old    0
63. Intraocular lens implantation in children is generally done at least a few years of age ( )    D. 5 years old    0
64. The main causes of childhood blindness in our country are ( )  A. infectious    0
64. The main causes of childhood blindness in our country are ( )  B. Nutritive  0
64. The main causes of childhood blindness in our country are ( )  C. Traumatic 0
64. The main causes of childhood blindness in our country are ( )  D. Congenital or hereditary    1
64. The main causes of childhood blindness in our country are ( )  E. Other 0
65. The refractive status of infants and young children has a change process with growth

and development, called ()    A. stabilization   0
65. The refractive status of infants and young children has a change process with growth and development, called ()    B. myopia    0
65. The refractive status of infants and young children has a change process with growth and development, called ()    C. farsightedness 0
65. The refractive status of infants and young children has a change process with growth and development, called ()    D. emmetropization   1
65. The refractive status of infants and young children has a change process with growth and development, called ()    E. Normalization 0
66. A test that provides a more accurate picture of a child's vision is () A.    Fixation reflex    0
66. A test that provides a more accurate picture of a child's vision is () B.    Follow reflection    0
66. A test that provides a more accurate picture of a child's vision is () C.    Priority viewing Method  1
66. A test that provides a more accurate picture of a child's vision is () D. Light sensing    0
67.The test that gives an overview of the child's vision is ()    A. Fixation reflex    1
67.The test that gives an overview of the child's vision is ()    B.    Priority viewing Method  0
67.The test that gives an overview of the child's vision is ()    C.    Visual evoked potential    0
67.The test that gives an overview of the child's vision is ()    D. Light sensing  0
68.The earliest age at which IOL implantation is generally performed in children is currently considered ()    A. 3 years    0
68.The earliest age at which IOL implantation is generally performed in children is currently considered ()    B. 6 months  0
68.The earliest age at which IOL implantation is generally performed in children is currently considered ()    C. 1 year    0
68.The earliest age at which IOL implantation is generally performed in children is currently considered ()    D. 2 years    1
68.The earliest age at which IOL implantation is generally performed in children is currently considered ()    E. 5 years   0
69. The refractive status of infants and young children has A process of change with growth and development, called()  A. Stabilization   0
69. The refractive status of infants and young children has A process of change with growth and development, called()  B. Myopic    0
69. The refractive status of infants and young children has A process of change with growth and development, called()  C. hypermetropia 0
69. The refractive status of infants and young children has A process of change with growth and development, called()  D. emmetropization    1
69. The refractive status of infants and young children has A process of change with growth and development, called()  E. Normalization 0
70. The following statements about amblyopia are true: ()    A.    optometry   is   not

necessary for children between 2 and 4 years of age   1
70. The following statements about amblyopia are true: ()   B. Optometry is not necessary for children before 4 years of age 0
70. The following statements about amblyopia are true: ()   C. Abnormal pupillary reflexia can also be a symptom of amblyopia   0
70. The following statements about amblyopia are true: ()   D. Children over 8 years of age do not respond to the treatment of amblyopia 0
70. The following statements about amblyopia are true: ()   E When covering, use full covering method  0
71. Visual function tests in children with amblyopia do not include ()  A. Fundus examination   0
71. Visual function tests in children with amblyopia do not include ()  B. Visual motor nystagmus   1
71. Visual function tests in children with amblyopia do not include ()  C. Priority viewing Method  0
71. Visual function tests in children with amblyopia do not include ()  D. Visual electrophysiological examination   0
71. Visual function tests in children with amblyopia do not include ()  E. Graphic eye chart   0
72. Amblyopia does not include which of the following ()   A. Form deprivation amblyopia   0
72. Amblyopia does not include which of the following ()   B. Anisometropic amblyopia   0
72. Amblyopia does not include which of the following ()   C. Ametropic amblyopia   0
72. Amblyopia does not include which of the following ()   D. Strabismatic amblyopia   0
72. Amblyopia does not include which of the following ()   E. Neonatal retinopathic amblyopia   1
Childhood retinopathy
73. More common in children ()   A. Central retinal artery obstruction   0
73. More common in children ()   B. Central serous chorioretinopathy   0
73. More common in children ()   C. oats disease   1
73. More common in children ()   D. Wet age-related macular degeneration   0
73. More common in children ()   E. retinopathy of prematurity  0
74. Causes of retinopathy of prematurity()  A. deformities that occur before birth   0
74. Causes of retinopathy of prematurity()  B. Sharp changes in blood partial oxygen pressure at birth  0
74. Causes of retinopathy of prematurity()  C. Inadequate oxygen delivery after birth   0
74. Causes of retinopathy of prematurity()  D. Excessive oxygenation after birth   1
74. Causes of retinopathy of prematurity()  E. Due to birth injury 0
75. Preterm infants with a history of prolonged oxygen use ()A. Central retinal artery obstruction   0

75. Preterm infants with a history of prolonged oxygen use ( ) B. Central serous chorioretinopathy    0
75. Preterm infants with a history of prolonged oxygen use ( ) C. oats disease    0
75. Preterm infants with a history of prolonged oxygen use ( ) D. Wet age-related macular degeneration 0
75. Preterm infants with a history of prolonged oxygen use ( ) E. retinopathy of prematurity   1
76. Caesarean section at 32 weeks gestation, birth weight 1.5 kg, history of oxygen inhalation, about 3 months after birth parents inadvertently found that the child's left eye pupils white. The most likely diagnosis for this child is( ) A. retinoblastoma    0
76. Caesarean section at 32 weeks gestation, birth weight 1.5 kg, history of oxygen inhalation, about 3 months after birth parents inadvertently found that the child's left eye pupils white. The most likely diagnosis for this child is( ) B.Coats disease   0
76. Caesarean section at 32 weeks gestation, birth weight 1.5 kg, history of oxygen inhalation, about 3 months after birth parents inadvertently found that the child's left eye pupils white. The most likely diagnosis for this child is( ) C. Retinopathy of prematurity  1
76. Caesarean section at 32 weeks gestation, birth weight 1.5 kg, history of oxygen inhalation, about 3 months after birth parents inadvertently found that the child's left eye pupils white. The most likely diagnosis for this child is( ) D. Congenital cataract    0
76. Caesarean section at 32 weeks gestation, birth weight 1.5 kg, history of oxygen inhalation, about 3 months after birth parents inadvertently found that the child's left eye pupils white. The most likely diagnosis for this child is( ) E. corneal adhesion leukoplakia  0
77. The following statement is not true is ( ) A. Central retinal artery occlusion can be induced by orbital surgeryb.   0
77. The following statement is not true is ( ) B.Choroidal neovascularization is more common in age-related macular degeneration   0
77. The following statement is not true is ( ) C. Maternal plaque includes retinopathy  1
77. The following statement is not true is ( ) D of prematurity. Wet age-related macular degeneration is more likely to have poor vision  0
77. The following statement is not true is ( ) E. Central serous chorioretinopathy is usually a self-limiting disease 0
78. Retinal neovascularization is rare in ( )   A. Ischemic retinal vein occlusion 0
78. Retinal neovascularization is rare in ( )   B. After acute onset of angle-closure glaucoma   1
78. Retinal neovascularization is rare in ( )   C. Diabetic retinopathy    0
78. Retinal neovascularization is rare in ( )   D. Preterm infants after oxygen    0
79. The child, male, 2 months, was born to the mother at 34 weeks of gestation, with a long birth Oxygen history, fundus is shown, the following is wrong about the disease ( )
    A. The fundus of this disease is characterized by fibrotic proliferation of the unvascularized retina 0

79. The child, male, 2 months, was born to the mother at 34 weeks of gestation, with a long birth Oxygen history, fundus is shown, the following is wrong about the disease ()
  B. Can cause retinal detachment, blindness 0
79. The child, male, 2 months, was born to the mother at 34 weeks of gestation, with a long birth Oxygen history, fundus is shown, the following is wrong about the disease ()
  C. Laser or cryotherapy is feasible for stage 2 to 3 patients, and stage 4 to 5 patients Hyaline surgery 0
79. The child, male, 2 months, was born to the mother at 34 weeks of gestation, with a long birth Oxygen history, fundus is shown, the following is wrong about the disease ()
  D. It can occur in all newborns 1
79. The child, male, 2 months, was born to the mother at 34 weeks of gestation, with a long birth Oxygen history, fundus is shown, the following is wrong about the disease ()
  E. Also known as Terry syndrome 0
80. Coats' disease is not distinguished from () A. Retinoblastoma 0
80. Coats' disease is not distinguished from () B. Retinopathy of prematurity 0
80. Coats' disease is not distinguished from () C. Familial exudative vitreoretinopathy 0
80. Coats' disease is not distinguished from () D. Congenital microeyeball 1
81. Treatment of retinopathy of prematurity includes () A. Laser photocoagulation 0
81. Treatment of retinopathy of prematurity includes () B. Freeze 0
81. Treatment of retinopathy of prematurity includes () C. Vitreous surgery 0
81. Treatment of retinopathy of prematurity includes () D. These are all 1

Strabismus in children

82. The child has an esotropia, marked by limited outward rotation of the eye and narrowing of the eyelid when attempting inward rotation. The most likely diagnosis is()
  A. abducens nerve palsy 0
82. The child has an esotropia, marked by limited outward rotation of the eye and narrowing of the eyelid when attempting inward rotation. The most likely diagnosis is()
  B. Duane retrogressive ocular syndrome 1
82. The child has an esotropia, marked by limited outward rotation of the eye and narrowing of the eyelid when attempting inward rotation. The most likely diagnosis is()
  C. concomitant esotropia 0
82. The child has an esotropia, marked by limited outward rotation of the eye and narrowing of the eyelid when attempting inward rotation. The most likely diagnosis is()
  D. fixed esotropia 0
83. Child with paralytic strabismus, male, age 6. Diplopia examination: the separation of the lower left complex image is the largest, and the surrounding image is the left eye, which external eye muscle paralysis should be considered () A. Right superior rectus 0
83. Child with paralytic strabismus, male, age 6. Diplopia examination: the separation of the lower left complex image is the largest, and the surrounding image is the left eye, which external eye muscle paralysis should be considered () B. right inferior rectus 0

83. Child with paralytic strabismus, male, age 6. Diplopia examination: the separation of the lower left complex image is the largest, and the surrounding image is the left eye, which external eye muscle paralysis should be considered ()  C. left superior oblique    0
83. Child with paralytic strabismus, male, age 6. Diplopia examination: the separation of the lower left complex image is the largest, and the surrounding image is the left eye, which external eye muscle paralysis should be considered ()  D. left inferior rectus 1
83. Child with paralytic strabismus, male, age 6. Diplopia examination: the separation of the lower left complex image is the largest, and the surrounding image is the left eye, which external eye muscle paralysis should be considered ()  E. Right superior oblique muscle  0
84. The transverse diameter of the cornea in normal infants is about ()A.10mm    0
84. The transverse diameter of the cornea in normal infants is about ()B.10.5mm   1
84. The transverse diameter of the cornea in normal infants is about ()C.11mm    0
84. The transverse diameter of the cornea in normal infants is about ()D.11.5mm   0
84. The transverse diameter of the cornea in normal infants is about ()E.12mm 0
85. A child, female, 4 years old, whose parents found that the left eye tilted upward when the eyes looked to the right, the eye position was normal when the eyes looked to the left, and the head tilt test was positive, and the diagnosis was considered as ()  A. Left superior rectus paralysis  0
85. A child, female, 4 years old, whose parents found that the left eye tilted upward when the eyes looked to the right, the eye position was normal when the eyes looked to the left, and the head tilt test was positive, and the diagnosis was considered as ()  B. Right superior rectus paralysis 0
85. A child, female, 4 years old, whose parents found that the left eye tilted upward when the eyes looked to the right, the eye position was normal when the eyes looked to the left, and the head tilt test was positive, and the diagnosis was considered as ()  C. Left superior oblique muscle paralysis  1
85. A child, female, 4 years old, whose parents found that the left eye tilted upward when the eyes looked to the right, the eye position was normal when the eyes looked to the left, and the head tilt test was positive, and the diagnosis was considered as ()  D. Right superior oblique paralysis    0
85. A child, female, 4 years old, whose parents found that the left eye tilted upward when the eyes looked to the right, the eye position was normal when the eyes looked to the left, and the head tilt test was positive, and the diagnosis was considered as ()  E. Paralysis of the left external direct muscle  0
86.Children often suffer from()    A cicatricial entropion    0
86.Children often suffer from()    B congenital entropion    1
86.Children often suffer from()    C mechanical entropion  0
86.Children often suffer from()    D cicatricial ectropion    0
86.Children often suffer from()    E paralytic ectropion  0
87. Age of onset of primary infantile esotropia()A 7 months  0
87. Age of onset of primary infantile esotropia()B 3 months  0
87. Age of onset of primary infantile esotropia()C 4 months  0

87. Age of onset of primary infantile esotropia()  D 6 months   1
87. Age of onset of primary infantile esotropia()  E 5 months   0
88. The age at which binocular vision development is completed in children is()   A 14 years old   0
88. The age at which binocular vision development is completed in children is()   B 6 years old    1
88. The age at which binocular vision development is completed in children is()   C 10 years old   0
88. The age at which binocular vision development is completed in children is()   D 3 years old    0
88. The age at which binocular vision development is completed in children is()   E 12 years old   0
89. Pediatric strabismus with amblyopia()   A concomitant esotropia   1
89. Pediatric strabismus with amblyopia()   B paralytic esotropia   0
89. Pediatric strabismus with amblyopia()   C poor visual acuity at one eye, good visual acuity at one eye   0
89. Pediatric strabismus with amblyopia()   D good visual acuity in both eyes   0
89. Pediatric strabismus with amblyopia()   E all more than correct   0
90. Non-objective test methods for visual acuity examination of children under 3 years of age include()   A observation of pupil response to light   0
90. Non-objective test methods for visual acuity examination of children under 3 years of age include()   B common visual acuity chart  1
90. Non-objective test methods for visual acuity examination of children under 3 years of age include()   C two-eye fixation response   0
90. Non-objective test methods for visual acuity examination of children under 3 years of age include()   D electrophysiological examination   0
90. Non-objective test methods for visual acuity examination of children under 3 years of age include()   E covering method   0
91. When treating intermittent alternating esotropia in children, it is first necessary to investigate whether there is()   A onset time  0
91. When treating intermittent alternating esotropia in children, it is first necessary to investigate whether there is()   B myopia   0
91. When treating intermittent alternating esotropia in children, it is first necessary to investigate whether there is()   C abducent nerve palsy   0
91. When treating intermittent alternating esotropia in children, it is first necessary to investigate whether there is()   D amblyopia 0
91. When treating intermittent alternating esotropia in children, it is first necessary to investigate whether there is()   E farsightedness  1
Ocular blastoma in children
92. Coats disease should not be distinguished from the following diseases()   A. Retinoblastoma   0
92. Coats disease should not be distinguished from the following diseases()   B. Retinopathy of prematurity   0
92. Coats disease should not be distinguished from the following diseases()   C.

Familial exudative vitreoretinopathy    0

92. Coats disease should not be distinguished from the following diseases()    D. Congenital microeyeball   1

93. Patient, male, 2 years old, parents found white pupil of right eye for 3 months. Examination of retinal detachment in the right eye fundus, a yellowish white mass in the temporal side, with new blood vessels on the surface. Which of the following diseases do you consider ()    A. Retinoblastoma    1

93. Patient, male, 2 years old, parents found white pupil of right eye for 3 months. Examination of retinal detachment in the right eye fundus, a yellowish white mass in the temporal side, with new blood vessels on the surface. Which of the following diseases do you consider ()    B. Retinitis pigmentosa    0

93. Patient, male, 2 years old, parents found white pupil of right eye for 3 months. Examination of retinal detachment in the right eye fundus, a yellowish white mass in the temporal side, with new blood vessels on the surface. Which of the following diseases do you consider ()    C. Choroidal melanoma    0

93. Patient, male, 2 years old, parents found white pupil of right eye for 3 months. Examination of retinal detachment in the right eye fundus, a yellowish white mass in the temporal side, with new blood vessels on the surface. Which of the following diseases do you consider ()    D. Xanthoma    0

93. Patient, male, 2 years old, parents found white pupil of right eye for 3 months. Examination of retinal detachment in the right eye fundus, a yellowish white mass in the temporal side, with new blood vessels on the surface. Which of the following diseases do you consider ()    E. Retinal detachment    0

94. Which of the following is the most common primary intraocular malignancy in children ()    A. Schwannoma   0

94. Which of the following is the most common primary intraocular malignancy in children ()    B. Rhabdomyosarcoma    0

94. Which of the following is the most common primary intraocular malignancy in children ()    C. Choroidal melanoma    0

94. Which of the following is the most common primary intraocular malignancy in children ()    D. Retinoblastoma    1

94. Which of the following is the most common primary intraocular malignancy in children ()    E. Meningioma   0

95. Patient, male, 2 years old, full term vaginal delivery, no history of oxygen use. Parent found right White pupil, insensitive to light. Ultrasound indicates intraocular space occupying disease Variable, diameter about 6mm. No systemic disease, no family history of eye disease. What kind of disease may it be ()    A.    Metastatic endophthalmitis   0

95. Patient, male, 2 years old, full term vaginal delivery, no history of oxygen use. Parent found right White pupil, insensitive to light. Ultrasound indicates intraocular space occupying disease Variable, diameter about 6mm. No systemic disease, no family history of eye disease. What kind of disease may it be ()    B.    Retinopathy of prematurity   0

95. Patient, male, 2 years old, full term vaginal delivery, no history of oxygen use.

Parent found right White pupil, insensitive to light. Ultrasound indicates intraocular space occupying disease Variable, diameter about 6mm. No systemic disease, no family history of eye disease.What kind of disease may it be ()  C. Retinoblastoma     1

95. Patient, male, 2 years old, full term vaginal delivery, no history of oxygen use. Parent found right White pupil, insensitive to light. Ultrasound indicates intraocular space occupying disease Variable, diameter about 6mm. No systemic disease, no family history of eye disease.What kind of disease may it be ()  D. Congenital cataract     0

95. Patient, male, 2 years old, full term vaginal delivery, no history of oxygen use. Parent found right White pupil, insensitive to light. Ultrasound indicates intraocular space occupying disease Variable, diameter about 6mm. No systemic disease, no family history of eye disease.What kind of disease may it be ()  E. Exudative retinopathy  0

96.Which treatment is the best choice for retinoblastoma treatment ()  A.    Follow-up observation  0

96.Which treatment is the best choice for retinoblastoma treatment ()  B.      Local radiotherapy 1

96.Which treatment is the best choice for retinoblastoma treatment ()  C.  condensation     0

96.Which treatment is the best choice for retinoblastoma treatment ()  D. chemotherapy    0

96.Which treatment is the best choice for retinoblastoma treatment ()  E.    enucleation     0

97. The following descriptions of retinoblastoma are wrong()A.   Binocular  disease accounts for about 30% to 35%    0

97. The following descriptions of retinoblastoma are wrong()B.   Calcification  can occur   0

97. The following descriptions of retinoblastoma are wrong()C. It is the most common primary intraocular malignant tumor in children 0

97. The following descriptions of retinoblastoma are wrong()D 90% occur before the age of 3 0

97. The following descriptions of retinoblastoma are wrong()E, more men than women    1

98.Which of the following is the most common primary intraocular malignancy in children()    A. schwannoma  0

98.Which of the following is the most common primary intraocular malignancy in children()    B. Rhabdomyosarcoma    0

98.Which of the following is the most common primary intraocular malignancy in children()    C. Choroidal melanoma   0

98.Which of the following is the most common primary intraocular malignancy in children()    D. retinoblastoma     1

98.Which of the following is the most common primary intraocular malignancy in children()    E. Meningioma   0

99. The following statement is wrong()A.  Most  of  the  holes  in  hiatus  retinal detachment are located in the superior temporal quadrant     0

99. The following statement is wrong()B. The superior temporal branch of retinal vein

occlusion is the most common 0

99. The following statement is wrong () C. Laser photocoagulation can treat central serous chorioretinopathy  0

99. The following statement is wrong () D. Retinoblastoma is more common in infants and young children  0

99. The following statement is wrong () E. Decreased vision is the earliest symptom of retinitis pigmentosa  1

100. The patient, a 2-year-old male, was diagnosed with white pupil of the right eye for 3 months. Examination: Retinal detachment in the right eye fundus, a yellowish white mass in the temporal side, with neovascularization on the surface. Which of the following diseases do you consider ()  A. Retinoblastoma  1

100. The patient, a 2-year-old male, was diagnosed with white pupil of the right eye for 3 months. Examination: Retinal detachment in the right eye fundus, a yellowish white mass in the temporal side, with neovascularization on the surface. Which of the following diseases do you consider ()  B. Retinitis pigmentosa  0

100. The patient, a 2-year-old male, was diagnosed with white pupil of the right eye for 3 months. Examination: Retinal detachment in the right eye fundus, a yellowish white mass in the temporal side, with neovascularization on the surface. Which of the following diseases do you consider ()  C. Choroidal melanoma  0

100. The patient, a 2-year-old male, was diagnosed with white pupil of the right eye for 3 months. Examination: Retinal detachment in the right eye fundus, a yellowish white mass in the temporal side, with neovascularization on the surface. Which of the following diseases do you consider ()  D. xanthoma 0

100. The patient, a 2-year-old male, was diagnosed with white pupil of the right eye for 3 months. Examination: Retinal detachment in the right eye fundus, a yellowish white mass in the temporal side, with neovascularization on the surface. Which of the following diseases do you consider ()  E. Retinal detachment  0